\documentclass[10pt,twocolumn,letterpaper]{article}

\usepackage{cvpr}              

\usepackage{graphicx}
\usepackage{amsmath}
\usepackage{amssymb}
\usepackage{booktabs}

\usepackage{algorithm}
\usepackage{algorithmic}
\usepackage{graphicx}
\usepackage{multirow}
\usepackage{tabularx}
\usepackage{soul}

\usepackage[accsupp]{axessibility}  

\newcommand{\crc}[1]{{#1}}
\newcommand{\wc}[1]{#1}

\usepackage[pagebackref,breaklinks,colorlinks]{hyperref}

\usepackage[capitalize]{cleveref}
\crefname{section}{Sec.}{Secs.}
\Crefname{section}{Section}{Sections}
\Crefname{table}{Table}{Tables}
\crefname{table}{Tab.}{Tabs.}

\def\cvprPaperID{1683} 
\def\confName{CVPR}
\def\confYear{2023}

\begin{document}

\title{CF-Font: Content Fusion for Few-shot Font Generation}

\author {
    Chi Wang\textsuperscript{\rm 1,2}\footnotemark[1]\ ,
    Min Zhou\textsuperscript{\rm 2},
    Tiezheng Ge\textsuperscript{\rm 2},
    Yuning Jiang\textsuperscript{\rm 2},
    Hujun Bao\textsuperscript{\rm 1},
    Weiwei Xu\textsuperscript{\rm 1}\footnotemark[2] \\
    \textsuperscript{\rm 1} State~Key~Lab~of~CAD\&CG,~Zhejiang~University \quad     \textsuperscript{\rm 2} Alibaba~Group\\
    {\tt\small wangchi1995@zju.edu.cn, \{yunqi.zm, tiezheng.gtz, mengzhu.jyn\}@alibaba-inc.com} \\
    {\tt\small \{bao, xww\}@cad.zju.edu.cn}
}

\twocolumn[{%
\renewcommand\twocolumn[1][]{#1}%
\maketitle
\begin{center}
    \centering
    \captionsetup{type=figure}
    \includegraphics[width=1.\textwidth]{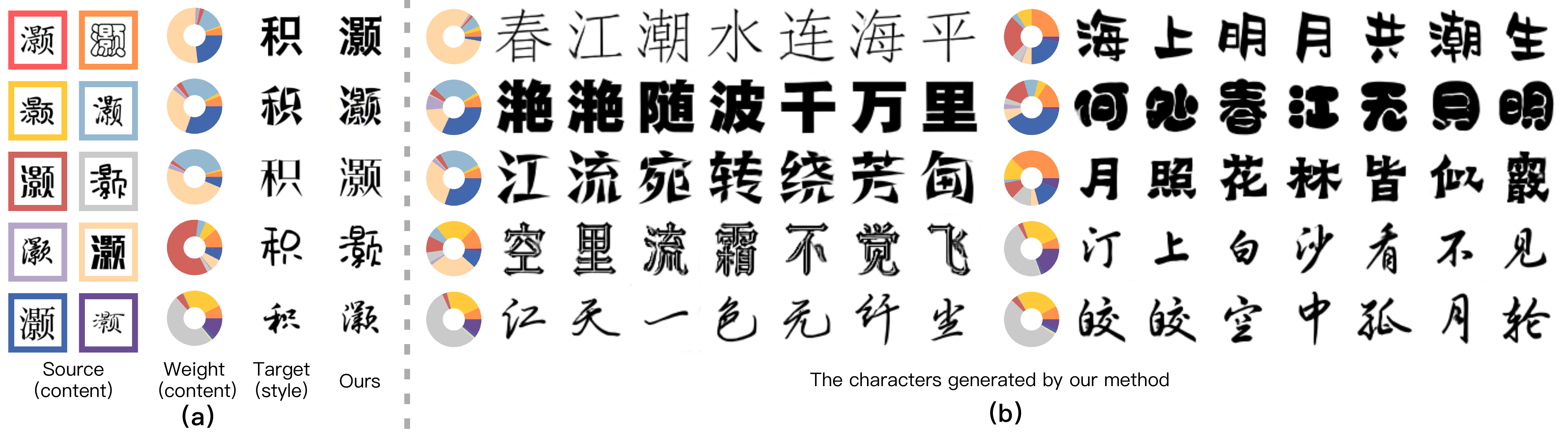}
    \captionof{figure}{\wc{Characters generated by our method.  (a) Source: source character images selected from ten basis fonts for content feature fusion. Weights: different colors and their covered areas on the doughnut chart represent the weights used to blend content features adaptively. Ten colors correspond to source images in colored boxes. 
    Target: few-shot target reference character images. One of those is performed as an example. 
    Ours: images generated by our method with fused content features and style features. (b) Generated character images of the first ten lines from a famous Chinese poem, each line with an extracted style, \eg thin, thick, swollen, cuneiform, inscription, or cursive style.}}
    \label{fig:teaser}
\end{center}%
}]

{
  \renewcommand{\thefootnote}{\fnsymbol{footnote}}
  \footnotetext[1]{This work was done during an internship at Alibaba Group.}
  \footnotetext[2]{Corresponding author.}
}

\begin{abstract}
Content and style disentanglement is an effective way to achieve few-shot font generation. It allows to transfer the style of the font image in a source domain to the style defined with a few reference images in a target domain. However, the content feature extracted using a representative font might not be optimal. In light of this, we propose a content fusion module~(CFM) to project the content feature into a linear space defined by the content features of basis fonts, which can take the variation of content features caused by different fonts into consideration. Our method also allows to optimize the style representation vector of reference images through a lightweight  iterative style-vector refinement~(ISR) strategy. Moreover, we treat the 1D projection of a character image as a probability distribution and leverage the distance between two distributions as the reconstruction loss~(namely projected character loss, PCL). Compared to L2 or L1 reconstruction loss, the distribution distance pays more attention to the global shape of characters. We have evaluated our method on a dataset of 300 fonts with 6.5k characters each. Experimental results verify that our method outperforms existing state-of-the-art few-shot font generation methods by a large margin. The source code can be found at \url{https://github.com/wangchi95/CF-Font}.
\end{abstract}

\section{Introduction}
Few-shot font generation aims to produce characters of a new font by transforming font images from a source domain to a target domain according to just a few reference images. It can greatly reduce the labor of expert designers to create a new style of fonts, especially for logographic languages that contain multiple characters, such as Chinese (over 60K characters), Japanese (over 50K characters), and Korean (over 11K characters), since only several reference images need to be manually designed. Therefore, font generation has wide applications in font completion for ancient books and monuments, personal font generation, etc. 

Recently, with the rapid development of convolutional neural networks~\cite{cnn} and generative adversarial networks~\cite{DBLP:conf/nips/GoodfellowPMXWOCB14}~(GAN), pioneers have made great progress in generating gratifying logographic fonts. Zi2zi~\cite{Zi2zi} introduces pix2pix~\cite{DBLP:conf/cvpr/IsolaZZE17} method to generate complex characters of logographic languages with high quality, but it cannot handle those fonts that do not appear in training~(unseen fonts). For the few-shot font generation, many methods~\cite{emd_cvpr18, savae_ijcai, gao2020gan, dmfont_eccv20, lffont_aaai21, mxfont_iccv21, DGFont_cvpr21} verify that content and style disentanglement is effective to convert the style of a character in the source domain, denoted as \emph{source character}, to the target style embodied with reference images of seen or unseen fonts. The neural networks in these methods usually have two branches to learn content and style features respectively, and the content features are usually obtained with the character image from a manually-chosen font, denoted as \emph{source font}. However, since it's a difficult task to achieve a complete disentanglement between content and style features~\cite{kwon2021diagonal, kazemi2019style}, the choice of the font for content-feature encoding influences the font generation results substantially. For instance, \emph{Song} and \emph{Kai} are commonly selected as the source font~\cite{lffont_aaai21, emd_cvpr18, DGFont_cvpr21, cggan_cvpr22, lyu2017auto,xu2009automatic}. While such choices are effective in many cases, the generated images sometimes contain artifacts, such as incomplete and unwanted strokes.

The main contribution of this paper is a novel content feature fusion scheme to \crc{mitigate the influence of incomplete disentanglement by exploring} the synchronization of content and style features, which significantly enhances the quality of few-shot font generation. \crc{Specifically, we design a content fusion module~(CFM) to take the content features of different fonts into consideration during training and inference.} It is realized by computing the content feature of a character of a target font through linearly blending content features of the corresponding characters in the automatically determined basis fonts, and the blending weights are determined through a carefully designed font-level distance measure. In this way, we can form a linear cluster for the content feature of a semantic character, and explore how to leverage the font-level similarity to seek for an optimized content feature in this cluster to improve the quality of generated characters.

In addition, we introduce an iterative style-vector refinement~(ISR) strategy to find a better style feature vector for font-level style representation. For each font, we average the style vectors of reference images and treat it as a learnable parameter. Afterward, we fine-tune the style vector with a reconstruction loss, which further improves the quality of the generated fonts. 

Most font-generation algorithms~\cite{Zi2zi,DGFont_cvpr21,lffont_aaai21, dmfont_eccv20, mxfont_iccv21, cggan_cvpr22} choose L1 loss as the character image reconstruction loss. However, L1 or L2 loss mainly supervises per-pixel accuracy and is easily disturbed by the local misalignment of details. Hence, we employ a distribution-based projected character loss~(PCL) to measure the shape difference between characters. Specifically, by treating the 1D projection of 2D character images as a 1D probability distribution, PCL computes the distribution distance to pay more attention to the global properties of character shapes, resulting in the large improvement of skeleton topology transfer results. 

\wc{The CFM can be embedded into the few-shot font generation task to enhance the quality of generated results.} Extensive experiments verify that our method, \wc{referred to as CF-Font}, remarkably outperforms state-of-the-art methods on both seen and unseen fonts.
\wc{Fig.~\ref{fig:teaser} reveals that our method can generate high-quality fonts of various styles.}
\section{Related Works}
\subsection{Image-to-image Translation}
Image-to-image translation is the task of converting a source image to the target domain of reference images. Early methods~\cite{DBLP:conf/cvpr/IsolaZZE17, DBLP:conf/cvpr/Wang0ZTKC18, DBLP:conf/iccv/ZhuPIE17,DBLP:conf/cvpr/ShrivastavaPTSW17, DBLP:conf/cvpr/ChoiCKH0C18} utilize GAN~\cite{DBLP:conf/nips/GoodfellowPMXWOCB14} and yield vivid images. But they could only convert the source image to some specific domains (or categories), which is more limited in practical applications. Recently, some few-shot methods~\cite{DBLP:conf/iccv/0001HMKALK19,DBLP:conf/iccv/HuangB17,DBLP:journals/tip/ChenXYST19,DBLP:conf/iccv/BaekCUYS21, DBLP:conf/nips/BenaimW17, DBLP:conf/icml/KimCKLK17} are proposed. These methods disentangle the content and style, and can convert the source image to arbitrary styles only if a few reference images are provided. Further, RG-UNIT~\cite{DBLP:conf/iccv/Gu0H0021} proposes an image retrieval strategy to help domain transfer, \ie it finds images similar to the source in content but in the target domain, and extracts their content features as assistance. Though the retrieval strategy helps to generate more realistic images, it cannot be directly applied to font generation tasks. Because the retrieved image may still differ significantly from the target in content, as fonts are highly fine-grained. Thus, we build basis fonts and use fused content features to narrow the gap between the source and target domains.

\begin{figure*}[!t]
    \centering
    \resizebox{0.9\linewidth}{!}{
 	\fbox{\includegraphics[width=\textwidth]{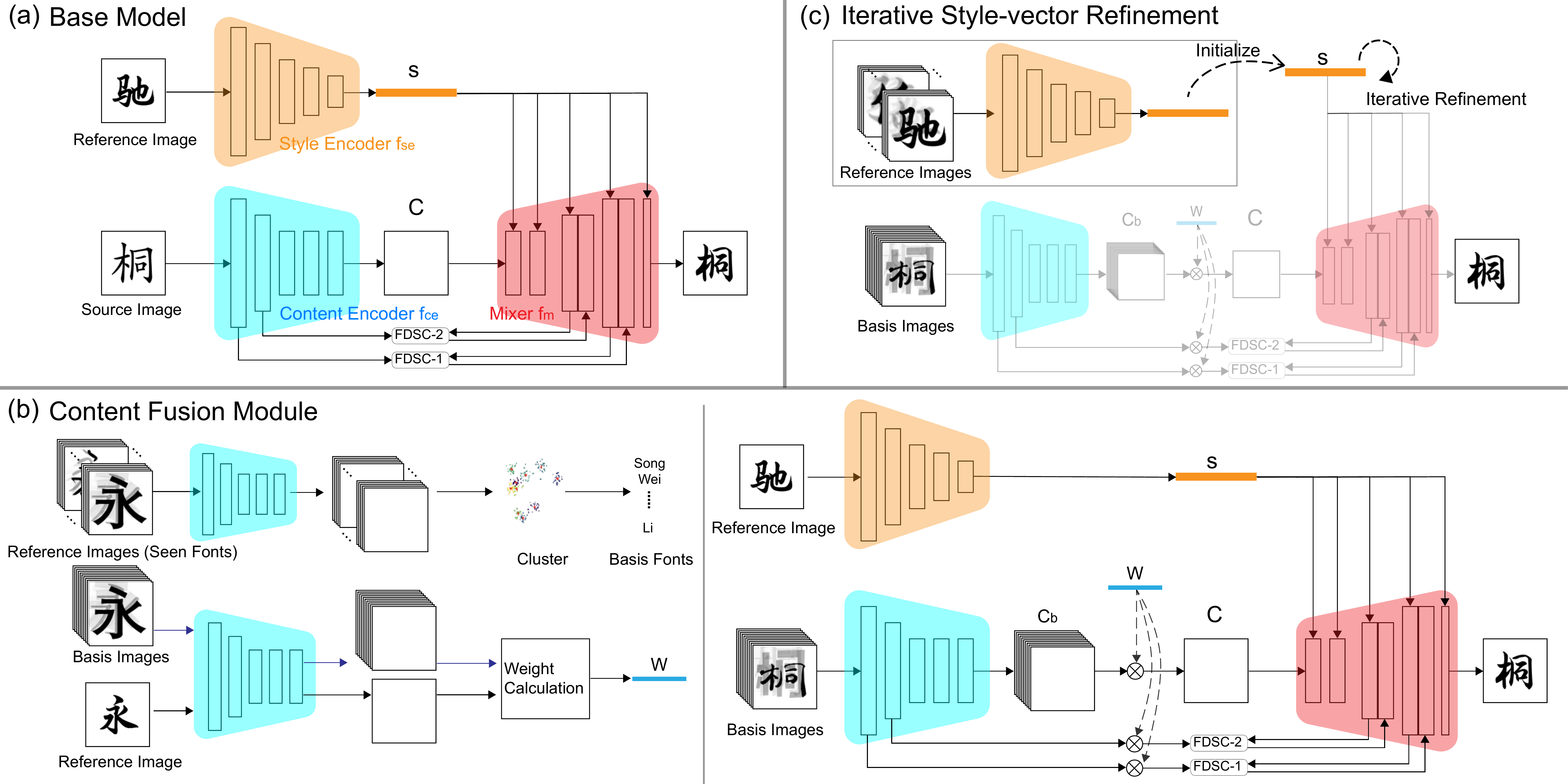}}
 	}
 	\vspace{-2mm}
    \caption{The framework of our model. (a) We first train \wc{the DGN}~\cite{DGFont_cvpr21} and use PCL to enhance the supervision of character skeletons. (b) After the model converges, content features of all training fonts are clustered and basis fonts are selected according to cluster centers. The original content encoder is replaced by CFM, and original content features are changed to fused features of basis fonts. Then we continue to train the model so that it adapts to fused content features. (c) In inference, we utilize ISR to polish the style of a font. 
    \wc{The extracted mean style vector is}
    treated as the only \wc{trainable} variable to be fine-tuned for a few iterations.}
    \label{fig:flowchart}
 	\vspace{-5mm}
\end{figure*}

\subsection{Few-shot Font Generation}
Few-shot font generation aims to generate a new font library in the required style with only a few reference images. 
Early methods~\cite{Zi2zi, DBLP:conf/bmvc/ChangGZW18,  DBLP:conf/icdar/LyuBYZHL17, DBLP:conf/siggrapha/JiangLTX17, DBLP:conf/icpr/SunZY18} for font generation train a cross-domain translation network to model mapping from the source to the target domain.
These structures limit the model to generate unseen fonts.
To address this issue, SA-VAE~\cite{savae_ijcai} and EMD~\cite{emd_cvpr18} disentangle the representations of style and content, and can generate images of all style-content combinations. RD-GAN~\cite{DBLP:conf/aaai/0006W20}, SCFont~\cite{DBLP:conf/aaai/JiangLTX19}, CalliGAN~\cite{DBLP:journals/corr/abs-2005-12500}, and LF-Font~\cite{lffont_aaai21} follow this way and employ component annotations to boost the style representation in local regions. To be less dependent on explicit component annotations, MX-Font~\cite{mxfont_iccv21} utilizes multiple experts and bipartite matching, and XMP-Font~\cite{xmpfont_cvpr22} employs a cross-modality encoder, which is conditioned jointly on character images and stroke labels. \wc{CG-GAN~\cite{cggan_cvpr22} supervises a font generator to decouple content and style on component level through a component-aware module.} But these \wc{three} methods still require the labels of component categories. \wc{Fs-Font is proposed to learn fine-grained local styles from reference images, and the spatial correspondence between the content and reference images.~\cite{fsfont} However, it needs to select reference characters carefully to achieve high-quality generated results.} DG-Font~\cite{DGFont_cvpr21} introduces a feature deformation skip connection module and achieves excellent performance without any extra labels. However, it is difficult for these few-shot methods to generate new fonts if the source and target domains are very different, especially when the target font is unseen. Starting from this perspective, we propose the CFM to reduce the difficulty of domain transfer, and the PCL to enhance skeleton supervision.

\section{Approach}

Our method is illustrated in Fig.~\ref{fig:flowchart}. 
The whole training pipeline can be divided into \wc{two} stages. 
Firstly, we train \wc{the neural network in DG-Font~\cite{DGFont_cvpr21} as our base network, referred to as DGN.}
\wc{The network is used to learn basic, disentangled content and style features of character images in our dataset.}
Secondly, \wc{our content fusion module~(CFM) is plugged into the model after the content encoder. 
Afterward, we replace the original content feature with the output of CFM, a linear content-feature interpolation of automatically-selected basis fonts. 
Then, we fix the content encoder and continue to train style encoder, feature deformation skip connection~\cite{DGFont_cvpr21}~(FSDC) and mixer together for a few epochs.}
The projected character loss~(PCL) is used in training to supervise character skeletons. 
\wc{In addition, to further improve the generation quality, we utilize the iterative style-vector refinement~(ISR) strategy to polish the learned font-level style vector alone in inference.} \crc{The motivation for ISR is seeking for a single and high-quality font-level style vector to generate images for all characters of the font.
Specifically, for a font, we refine upon the average of the character style vector of all the given 16 characters in our few-shot setting. 
}

\subsection{Base Network} 
As illustrated in Fig.~\ref{fig:flowchart}~(a), given a content image $\boldsymbol{I}_c$ and a style image $\boldsymbol{I}_s$, the DGN synthesizes an image with the character of the content image and the font of the style image. This generative network consists of four parts: a style encoder $f_{se}$ to extract style latent vector $\boldsymbol{s}$, a content encoder $f_{ce}$ to obtain content feature map $\boldsymbol{C}$, a mixer $f_m$ to mix style and content representations with AdaIN~\cite{DBLP:conf/iccv/HuangB17}, and two FSDC modules.
During training, a multi-task discriminator, fed with generated characters and their ground-truth images, is applied to conduct discrimination for each style simultaneously.

Four losses are adopted: 1) image reconstruction loss $\mathcal{L}_{img}$ for domain-invariant features maintaining; 2) content consistent loss $\mathcal{L}_{cnt}$ to guarantee consistency between generated and input content images; 3) adversarial loss $\mathcal{L}_{adv}$ in hinge version~\cite{DBLP:journals/corr/LimY17,DBLP:conf/iclr/MiyatoKKY18,DBLP:conf/icml/ZhangGMO19} for realistic image generation; 4) deformation offset normalization $\mathcal{L}_{offset}$ to avoid excessive offsets in FDSC. More details are in \cite{DGFont_cvpr21}.

\begin{figure}[t]
  \begin{center}
    \fbox{\includegraphics[width=0.8\linewidth]{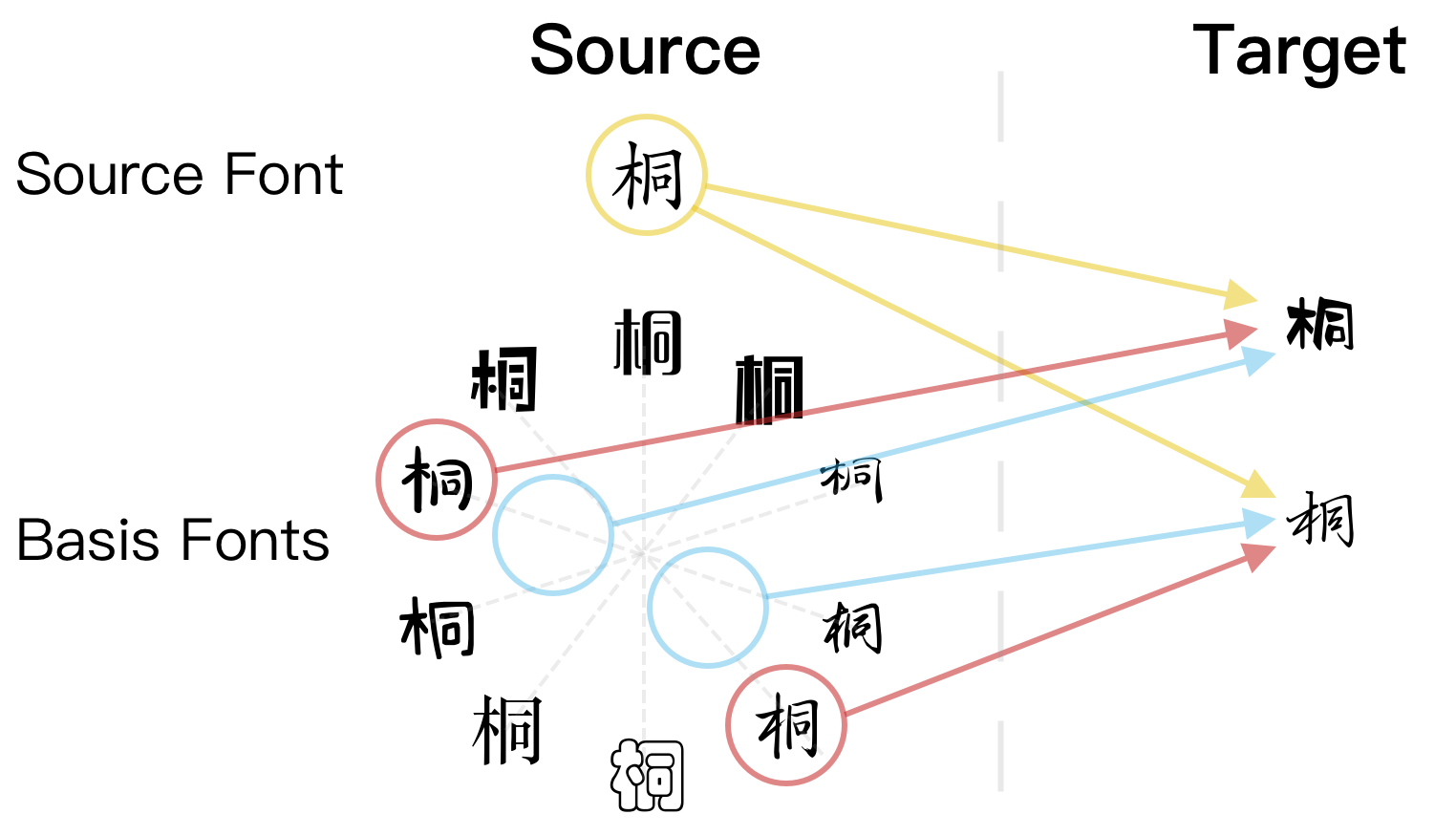}}
  \end{center}
  \caption{Visualization of content fusion. The yellow and red arrows are denoted for content features from the commonly used source font \emph{Kai}~\cite{lffont_aaai21} and the nearest font of the target respectively. The blue arrow represents the interpolation of content features of basis fonts to approximate the target.}
  \label{fig:single_vs_multi}
\end{figure}

\subsection{Content Fusion Module}
The content fusion module aims to adaptively extract content features by combining the content features of basis fonts. This network with CFM is constructed as in Fig.~\ref{fig:flowchart}~(b).
\wc{Firstly, to find representative fonts for content fusion, we cluster all training fonts through the concatenated content features of \crc{the given 16} few-shot characters and pick those nearest to the cluster centers as basis fonts. The basis fonts are fixed once selected.}
Then, for each target font, we calculate an L1-norm content fusion weight according to its similarity to basis fonts. 
As a result, the content features (input of the decoder) are replaced by the weighted sum of those of basis fonts. 
In addition, the network should be fine-tuned for a few epochs to adapt to fused content features~\wc{(represented as the blue circles in Fig.~\ref{fig:single_vs_multi})}.

\paragraph{Basis selection.} 
Suppose we need to choose $M$ basis fonts from $N$ training fonts. 
It can be realized by clustering the content features $\{\boldsymbol{C}_i\}_{i=1}^N$ and selecting fonts lying in the cluster centroids. 
In our practice, since the dimension of $\boldsymbol{C}_i$ is too large while $N$ is relatively small, we follow~\cite{thrun2021exploitation} to map $\boldsymbol{C}_i$ to the vector of the distances between it and features of all fonts $\boldsymbol{e}_i\in{\mathbb{R}^N}$. More specifically:
\begin{equation}
\label{eq:bs}
\begin{aligned}
      \boldsymbol{C}_i &= f_{ce}(\boldsymbol{I}_i),\footnotemark \\[1mm]
      \boldsymbol{d}_i &= (d_{i1}, d_{i2}, ..., d_{iN}), 
      \quad d_{ij} = \Vert \boldsymbol{C}_i-\boldsymbol{C}_j \Vert_1, \\[1mm]
      \quad \boldsymbol{e}_i &= \sigma(\boldsymbol{d}_i), \\[1mm]
      \mathcal{B} &= \mathbf{Cluster}(M, \{\boldsymbol{e}_1, \boldsymbol{e}_2, ..., \boldsymbol{e}_N\}), \\[1mm]
\end{aligned}
\end{equation}
\footnotetext{$\boldsymbol{C}_i$ is actually the \wc{concatenated} content features extracted from several characters of font $i$. For the sake of brevity, we omit the superscript for characters here.}where $\sigma(\cdot)$ is the softmax operation, $d_{ij}$ is the L1 distance between two fonts, $\mathbf{Cluster}$ is the classical K-Medoids cluster algorithm~\cite{k-meanspp06}, and set $\mathcal{B}$ is the indices of selected fonts.

\paragraph{Weight calculation.} For the target font $t$ and its content feature $\boldsymbol{C}_t$, we measure its similarity to the basis fonts $\{\boldsymbol{C}_m\}_{m=1}^M$, namely $\boldsymbol{d}_t^{\prime}\in{\mathbb{R}^M}$. Then the content fusion weight $\boldsymbol{w}_t\in{\mathbb{R}^M}$ is calculated as follow:
\begin{equation}
\label{eq:bw}
 \begin{aligned}
    \boldsymbol{d}_t^{\prime} &= (d_{t1}, d_{t2}, ..., d_{tM}), 
    \quad d_{tm} = \Vert \boldsymbol{C}_t-\boldsymbol{C}_m \Vert_1, \\[1mm]
    \boldsymbol{w}_t &= \sigma(-\boldsymbol{d}_t^{\prime} / \tau),
\end{aligned}
\end{equation}
where $\tau$ is the temperature of the softmax operation.

\paragraph{Content fusion.} Once the basis fonts and content fusion weights are obtained, \wc{the} original content feature map $\boldsymbol{C}$ is replaced with the fused one $\boldsymbol{C}_t^{\prime}$,
\wc{where the content fusion weight of CFM is related to its target font $t$.}
\begin{equation}
\label{eq:cf}
\begin{array}{rl}
    \boldsymbol{C}_t^{\prime} = \sum_{m \in \mathcal{B}}{w_{tm} \cdot \boldsymbol{C}_m}.
\end{array}
\end{equation}

\subsection{Projected Character Loss}

\begin{figure}[t]
    \centering
    \includegraphics[width=0.5\linewidth]{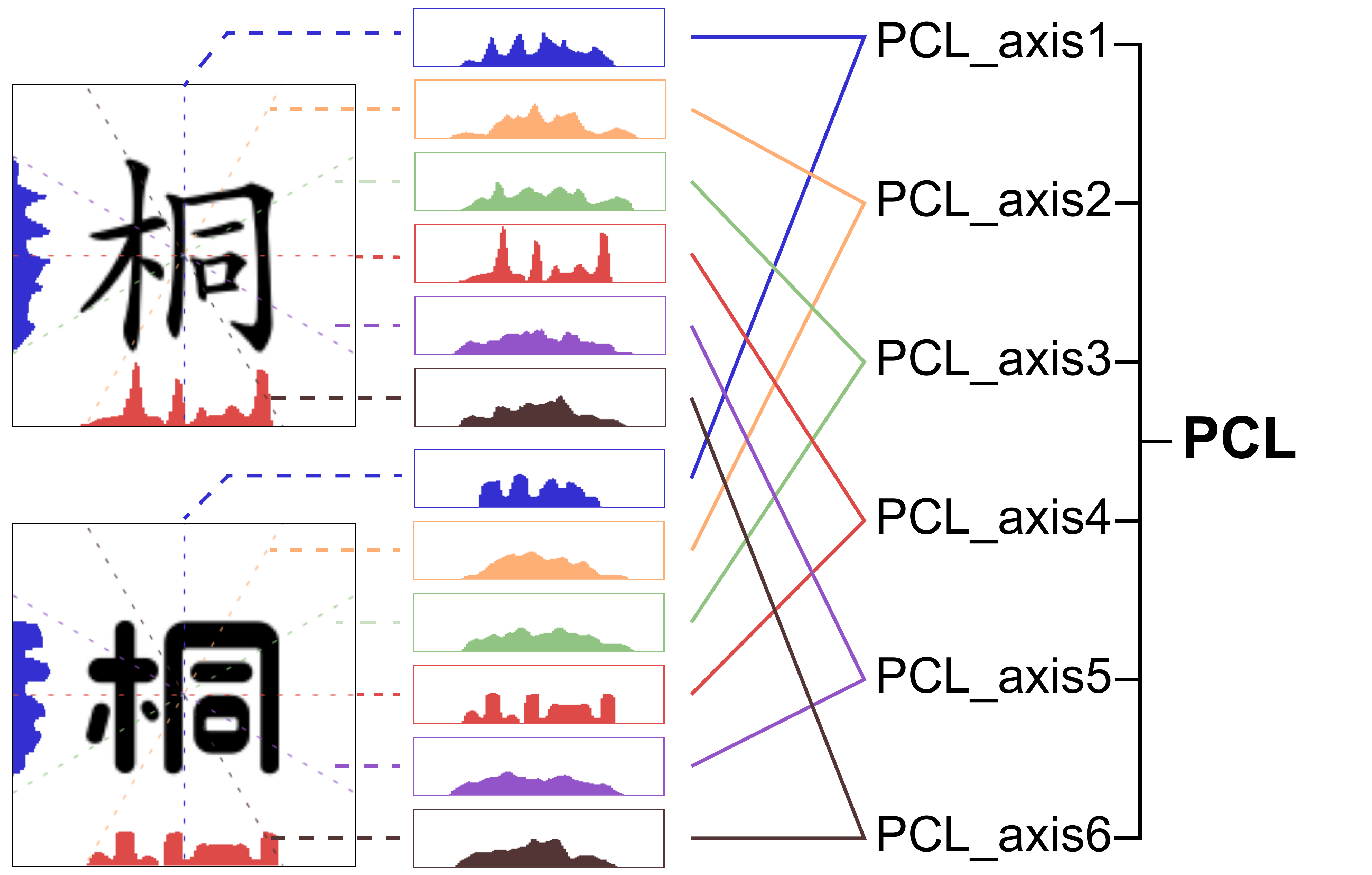}
    \caption{Illustration of PCL. We project the binary characters into multi-direction 1D spaces (distinguished by color) and calculate normalized histograms for each. It is obvious that for the different fonts with the same character, the projected distributions vary along with the skeletons and are less sensitive to textures or colors.}
    \label{fig:wdl}
 	\vspace{-2mm}
\end{figure}

To better supervise the skeleton, we design a projected character loss, which measures image difference with marginal distribution distances on multiple 1D projections \crc{, as shown in Fig.~\ref{fig:wdl}}. Since the distribution is sensitive to the relative relationship, PCL pays more attention to the global shape of characters.
\begin{equation}
\label{eq:pcl}
\mathcal{L}_p(\boldsymbol{Y}, \hat{\boldsymbol{Y}}) = \frac{1}{P}\sum^P_{p=1} \mathcal{L}_{1d}(\phi_p(\boldsymbol{Y}), \phi_p(\hat{\boldsymbol{Y}})), 
\end{equation}
where $\boldsymbol{Y}$ and $\hat{\boldsymbol{Y}}$ represent the generated and ground-truth image respectively, $P$ is the number of projections, and $\phi_p(\cdot)$ denotes a projection function with the $p$-th direction.

There are lots of metrics to measure the alignment between 1D distributions, such as the KL-divergence and Wasserstein distance. Thus, $\mathcal{L}_p$ can have various forms:
\begin{equation}
\label{eq:pwdl}
\begin{aligned}
    \mathcal{L}_{pc-wdl}(\boldsymbol{Y}, \hat{\boldsymbol{Y}}) &= \frac{1}{P}\sum^P_{p=1}  \left\|\frac{\Lambda(\phi_p(\boldsymbol{Y}))}{\sum \phi_p(\boldsymbol{Y})}-\frac{\Lambda(\phi_p(\hat{\boldsymbol{Y}}))}{\sum \phi_p(\hat{\boldsymbol{Y}})}\right\| \\
    \mathcal{L}_{pc-kl}(\boldsymbol{Y}, \hat{\boldsymbol{Y}}) &= \frac{1}{P}\sum^P_{p=1} \mathbf{KL}\Big( \frac{\phi_p(\boldsymbol{Y})}{\sum \phi_p(\boldsymbol{Y})},\frac{\phi_p(\hat{\boldsymbol{Y}})}{\sum \phi_p(\hat{\boldsymbol{Y}})}\Big),
\end{aligned}
\end{equation}
where $\mathbf{KL}$ means the KL-divergence and $\Lambda$ denotes the cumsum function, which turns probability density functions to cumulative distribution functions. 

\begin{figure}[t]
    	\centering
    \includegraphics[width=0.65\linewidth]{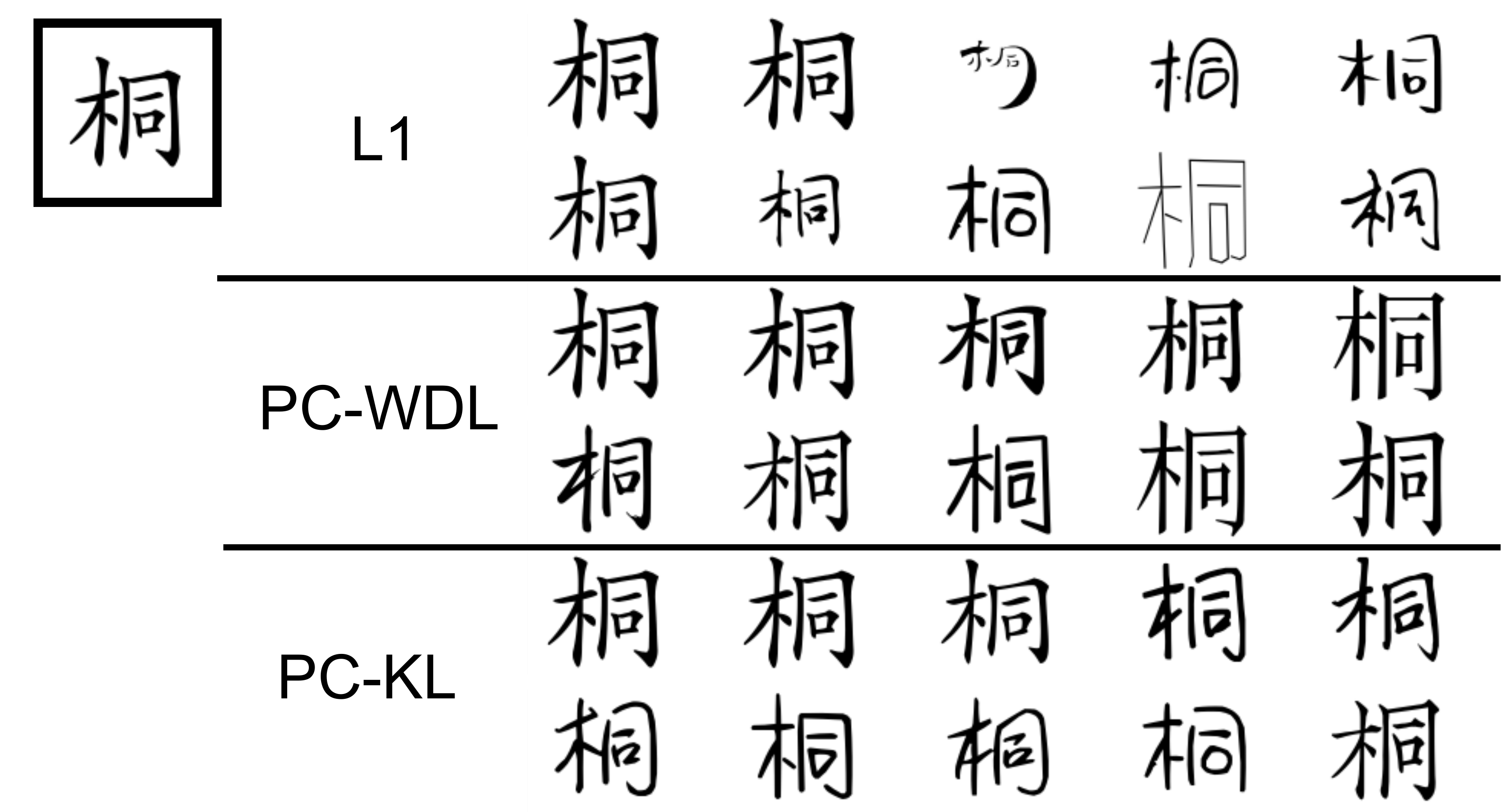}
    \caption{L1 vs PCL. We retrieve the closest character of all training fonts to the top-left one by L1, PC-WDL, and PC-KL, respectively. The top ten results of each loss are listed from left to right, top to down. It can be seen that the skeletons vary greatly in the column of L1 but are quite consistent in those of PCL.}
    \label{fig:l1vspcl}
\end{figure}

To simply verify the performance of PCL, we generate images of the character ``Tong'' from 240 fonts and measure their similarity by PCL and L1. The closest ten characters to the top-left one found by different metrics are displayed in Fig~\ref{fig:l1vspcl} respectively. It can be seen that the characters retrieved by L1 are quite different on the character skeleton, which is important for fonts. While those selected by PCL are relatively more consistent and it indicates that PCL is more proper for measuring the skeleton.

Adding PCL to the image reconstruction loss term, we have the following overall loss function for training:
\begin{equation}
\label{eq:pwdl2}
\begin{aligned}
\mathcal{L}=\mathcal{L}_{adv}+\lambda_{img} (\mathcal{L}_{img}+\lambda_{pcl}\mathcal{L}_{pcl})\\+\lambda_{cnt} \mathcal{L}_{cnt}+\lambda_{offset} \mathcal{L}_{offset},
\end{aligned}
\end{equation}
where $\lambda_{img}$, $\lambda_{pcl}$, $\lambda_{cnt}$, and $\lambda_{offset}$ are hyperparameters to adjust the weight of each loss function.

\subsection{Iterative Style-vector Refinement} 
For target font $t$, a robust style information can be extracted as the latent style vector $\boldsymbol{s}_t^{\prime}$ by averaging the outputs of $f_{se}$ with a set of character images~\cite{DGFont_cvpr21}.
\vspace{-5pt}
\begin{equation}
\label{eq:sii_init}
\boldsymbol{s}_t^{\prime} = \frac{1}{Q}{\sum_{q=1}^Q f_{se}(\boldsymbol{I}^q_t)},
\vspace{-5pt}
\end{equation}
where $\boldsymbol{I}^q_t$ is an image of character $q$ of font $t$, and Q denotes the reference character number.

Motivated by the "iterative inference" strategy that optimizes input in the inference stage~(\eg ~\cite{yeh2017semantic}), we propose iterative style-vector refinement for further optimizing the style feature $\boldsymbol{s}_t^{\prime}$. As in Fig.~\ref{fig:flowchart}~(c), in the inference stage, $\boldsymbol{s}_t^{\prime}$ is first initialized by Eq.~\ref{eq:sii_init}. Then, using the provided few \wc{reference} characters of target fonts $\{\boldsymbol{I}_t^{q}\}_{q=1}^Q$ as supervising samples, we refine $\boldsymbol{s}_t^{\prime}$ for around ten epochs according to the backpropagation of the reconstruction loss. Finally, the optimized style vector is adopted for inference. Worth noting this style vector can be stored as a signature of the target font and reused in referencing all characters of the same font, which makes the proposed ISR efficient in the real system.
\section{Experiments}

\newcommand{\tabincell}[2]{\begin{tabular}{@{}#1@{}}#2\end{tabular}}
\begin{table*}[t]
  \caption{Comparison with state-of-the-art methods on seen/unseen fonts. Bold and underlined numbers denote the best and the second best respectively. The numbers in the last row represent our improvement over the second-best scores.}
  \label{tbl:vs_sota}
  \vspace{-5pt}
  \centering
  \resizebox{0.75\linewidth}{!}{
  \begin{tabular}{l@{\hspace{0.07in}}c@{\hspace{0.07in}}c@{\hspace{0.07in}}c@{\hspace{0.07in}}c@{\hspace{0.07in}}c@{\hspace{0.07in}}c@{\hspace{0.07in}}c@{\hspace{0.07in}}c@{\hspace{0.07in}}c@{\hspace{0.07in}}c@{\hspace{0.07in}}c}
    \toprule
    \multirow{2}*{Methods} & \multicolumn{5}{c}{Seen Fonts} & \multicolumn{5}{c}{Unseen Fonts}&\multirow{2}*{User Study \%} \\ \cmidrule(r){2-6}\cmidrule(r){7-11}
         & L1$\downarrow$ & RMSE$\downarrow$  & SSIM $\uparrow$ & LPIPS$\downarrow$ & FID$\downarrow$  & L1$\downarrow$ & RMSE$\downarrow$  & SSIM $\uparrow$ & LPIPS$\downarrow$ & FID$\downarrow$  \\
    \midrule
    FUNIT & 0.08591 & 0.2529 & 0.6661 & 0.1169 & \textbf{11.66} & 0.09377 & 0.2686 & 0.6432 & 0.1427 & 28.10 & 11.74 \\
    LF-Font & 0.08098 & 0.2435 & 0.6829 & 0.1226 & 27.73 & 0.09037 & 0.2620 & 0.6534 & 0.1448 & 38.46 & 13.01 \\
    MX-Font & 0.07470 & 0.2319 & 0.7038 & 0.1034 & 18.75 & 0.08171 & 0.2468 & 0.6830 & \underline{0.1193} & \underline{27.91} & 10.86\\
    Fs-Font & 0.08214 & 0.2519 & 0.6657 & 0.1502 & 45.33 & 0.08917 & 0.2657 & 0.6467 & 0.1647 & 55.21 & 12.03 \\
    CG-GAN & 0.07977 & 0.2409 & 0.6883 & 0.1117 & 23.93 & 0.08639 & 0.2549 & 0.6690 & 0.1303 & 37.22 & 16.67 \\ 
    \midrule
    DG-Font & \underline{0.06251} & \underline{0.2105} & \underline{0.7437} & \underline{0.0846} & 17.10 & \underline{0.07841} & \underline{0.2442} & \underline{0.6853} & 0.1198 & 27.98 & 14.11\\
    \tabincell{l}{CF-Font\\ \ } & \tabincell{c}{\textbf{0.05997}\\(4.1\%)} & \tabincell{c}{\textbf{0.2053}\\(2.5\%)} & \tabincell{c}{\textbf{0.7538}\\(1.4\%)} & \tabincell{c}{\textbf{0.0836}\\(1.1\%)} & \tabincell{c}{\underline{13.13}\\(-)} & \tabincell{c}{\textbf{0.07394}\\(5.7\%)} & \tabincell{c}{\textbf{0.2354}\\(3.6\%)} & \tabincell{c}{\textbf{0.7007}\\(2.3\%)} & \tabincell{c}{\textbf{0.1182}\\(0.92\%)} & \tabincell{c}{\textbf{26.51}\\(5.0\%)} & \tabincell{c}{\textbf{21.58}\\(29.5\%)} \\ 
    \bottomrule
  \end{tabular}
  }
\end{table*}


\begin{figure*}[t]
	\centering
    \setlength{\tabcolsep}{1pt}
 	\resizebox{0.8\linewidth}{!}{
	\renewcommand\arraystretch{1.47}
	\begin{tabular}{l@{\hspace{0.3in}}l@{\hspace{-0.2in}}|l@{\hspace{0.0in}}}
        \hline
        \multirow{9}*{\rotatebox[origin=c]{90}{Seen Fonts}} & \makebox[0.11\linewidth][l]{Source} &   \multirow{9}*{\includegraphics[width=0.89\linewidth]{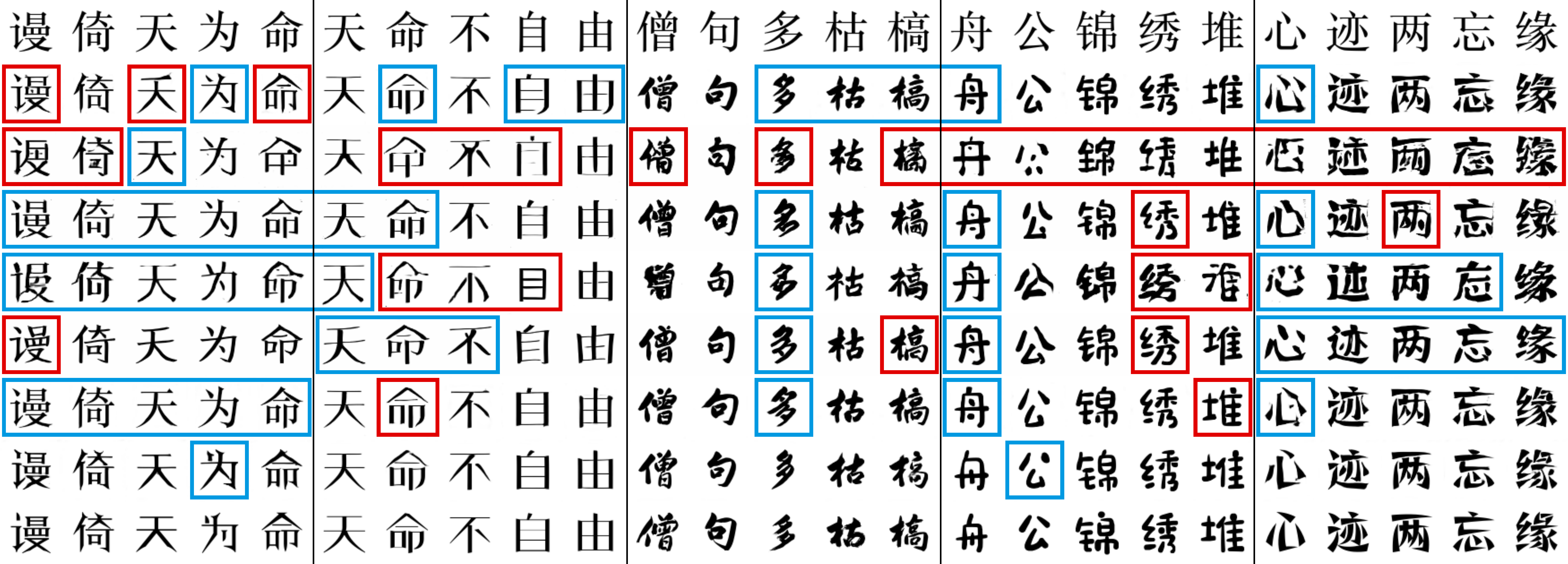}}\\
        &\makebox[0.11\linewidth][l]{FUNIT} & \\
		&\makebox[0.11\linewidth][l]{LF-Font} & \\
        &\makebox[0.11\linewidth][l]{MX-Font} & \\
        &\makebox[0.11\linewidth][l]{Fs-Font} & \\
        &\makebox[0.11\linewidth][l]{CG-GAN} & \\
        &\makebox[0.11\linewidth][l]{DG-Font} & \\
        &\makebox[0.11\linewidth][l]{CF-Font} & \\
        &\makebox[0.11\linewidth][l]{Target} & \\

        \hline
        \multirow{9}*{\rotatebox[origin=c]{90}{Unseen Fonts}} & \makebox[0.11\linewidth][l]{Source} &   \multirow{9}*{\includegraphics[width=0.89\linewidth]{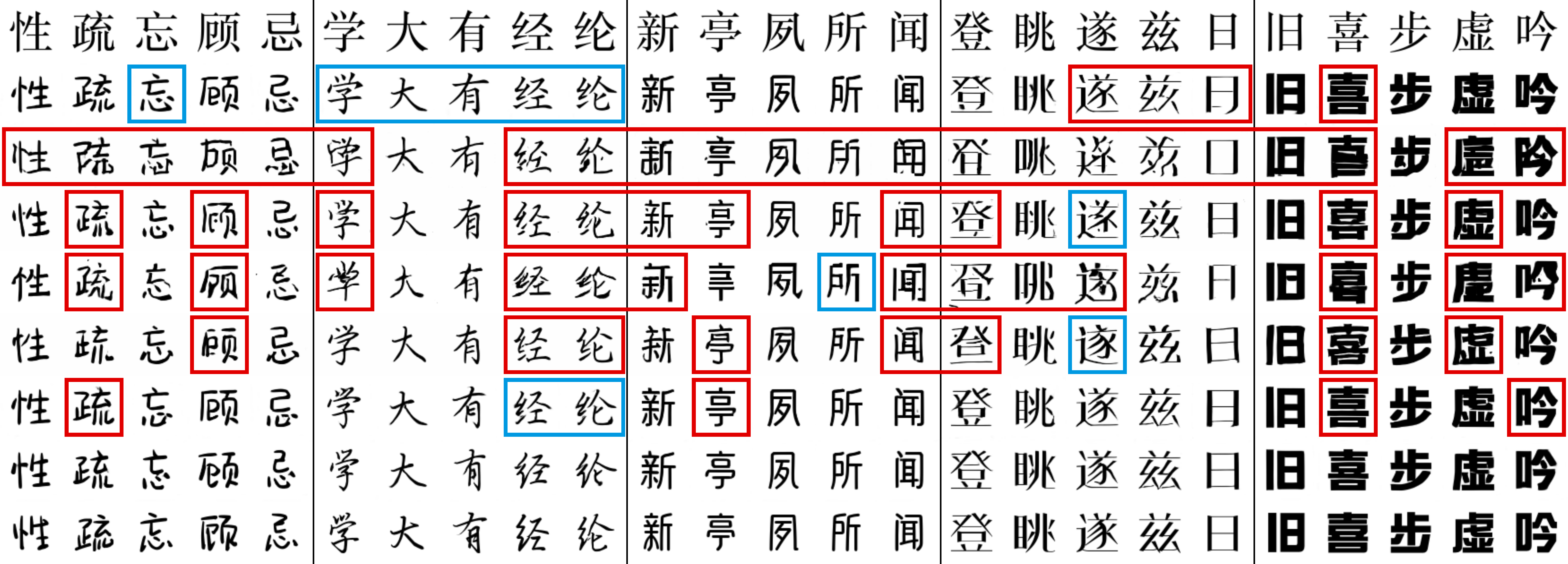}}\\
        &\makebox[0.11\linewidth][l]{FUNIT} & \\
		&\makebox[0.11\linewidth][l]{LF-Font} & \\
        &\makebox[0.11\linewidth][l]{MX-Font} & \\
        &\makebox[0.11\linewidth][l]{Fs-Font} & \\
        &\makebox[0.11\linewidth][l]{CG-GAN} & \\
        &\makebox[0.11\linewidth][l]{DG-Font} & \\
        &\makebox[0.11\linewidth][l]{CF-Font} & \\
        &\makebox[0.11\linewidth][l]{Target} & \\
        \hline
	\end{tabular}
	}
	\vspace{-2mm}
    \caption{Qualitative comparison with state-of-the-art methods \crc{on Chinese poems}. As mentioned earlier, we use multiple source fonts and pick the best results for these comparison methods for fairness. Here we just plot font \emph{Song} as an example of source fonts for convenience. \wc{We mark erroneous skeletons with red boxes and other mismatch styles, such as stroke style, joined-up style, and body frame~\cite{lu2016elements}, with blue boxes.}}
	\label{fig:vs_sota_big}
	\vspace{-10pt}
\end{figure*}

\wc{We have implemented the CF-Font method on a GPU server with 8 Nvidia Tesla V100 GPUs. After training with our dataset, our method outperforms the state-of-the-art methods on unseen fonts by 5.7\% and 5.0\% with respect to L1 and FID metrics, respectively. In the following, we report the preparation of dataset, evaluation metrics, and various experimental results to verify the effectiveness of our method.}
\subsection{Dataset and Evaluation Metrics}
\label{ssec:data}
We collect 300 Chinese fonts to build a dataset (including printed and handwriting fonts) to verify our method for the Chinese font generation task.  \wc{Our character set~(6446 in total) covers almost the full standard Chinese character set~(6763 in total) of GB/T 2312~\cite{CJKV}, and 317 characters not supported by comparison methods are removed.} The training part contains 240 fonts, and each font has 800 characters. The test part consists of (a) 229 seen fonts with 5646 unseen characters; (b) the remaining \wc{60} unseen fonts with \wc{5646} unseen characters, to verify the generalization ability of models. Note that we exclude 11 of the \wc{240} training fonts when testing on seen fonts. They are basis fonts (including \emph{Song}) in CFM and a font \emph{Kai}, in which \emph{Song} and \emph{Kai} are commonly used as source fonts in font generation~\cite{lffont_aaai21, emd_cvpr18, DGFont_cvpr21, cggan_cvpr22}. Besides, for few-shot font generation, reference images of target fonts in the test are with 16 randomly picked characters from the training part. 

We leverage pixel-level and perceptual metrics for evaluation, following \cite{DGFont_cvpr21}. Specifically, pixel-level metrics are L1, root mean square error~(RMSE), and structural similarity index measure~(SSIM). They focus on per-pixel consistency between generated images and ground-truth ones. Perceptual metrics include FID~\cite{fid_nips17} and LPIPS~\cite{lpips_cvpr18}, both of which measure the similarity of features and are closer to human vision.

\subsection{Implementation Details}
We train our model using Adam~\cite{DBLP:journals/corr/KingmaB14} with $\beta_1=0.9$ and $\beta_2=0.99$ for the style encoder, and RMSprop~\cite{RMSprop} with $\alpha=0.99$ for the content encoder. The learning rate and weight decay are both set as $10^{-4}$. The hyper-parameters for loss are $\lambda_{img}=\lambda_{cnt}=0.1$, $\lambda_{offset}=0.5$, and $\lambda_{pcl}=0.01$ ($0.05$ for PC-KL). For PCL, we orthographically project a character image onto 12 straight lines, which cross at the image center and divide the 2D space evenly. We resize all images to 80 × 80 and train the model with a batch size of 32. \wc{The whole training takes about 15 hours.} 
We first train the \wc{DGN} for 180k iterations to obtain the learned content features. 
Then we cluster these content features into ten groups and select basis fonts by the distance to cluster centers. 
\wc{After that, the model with CFM is further trained for another 20k iterations.}
For fairness, the models without CFM in ablations are trained for 200k iterations.

\subsection{Comparison with State-Of-The-Art Methods}
We compare our model with \wc{six} state-of-the-art methods, including an image-to-image translation method~(FUNIT\cite{DBLP:conf/iccv/0001HMKALK19}), four component-related methods~(LF-Font~\cite{lffont_aaai21}, MX-Font~\cite{mxfont_iccv21}, CG-GAN~\cite{cggan_cvpr22}, FsFont~\cite{fsfont}), and DG-Font~\cite{DGFont_cvpr21}. \wc{We slightly modify the network of CG-GAN to fit the input image size and the few-shot setting.} 
To be fair, we try each of our basis fonts and font \emph{Kai} as the source font for these comparison methods and report their best results in the following part (see details in our supplementary).

\begin{table*}[t]
  \caption{Ablation studies on different components. The first row is the result of DGN. P, C and S represent PC-WDL, CFM and ISR respectively. N means using retrieval strategy, \ie picking the closet font from basis fonts~(if marked with a star *, from the whole training set expect the target font itself) as the source according to the similarity between content features.}
  \label{tbl.wcs}
  \vspace{-5pt}
  \centering
  \resizebox{0.8\linewidth}{!}{
  \begin{tabular}{l@{\hspace{0.07in}}l@{\hspace{0.07in}}l@{\hspace{0.07in}}l@{\hspace{0.07in}}c@{\hspace{0.07in}}l@{\hspace{0.07in}}l@{\hspace{0.07in}}l@{\hspace{0.07in}}l@{\hspace{0.07in}}c@{\hspace{0.07in}}l@{\hspace{0.07in}}l@{\hspace{0.07in}}l@{\hspace{0.07in}}l}
    \toprule     \multicolumn{4}{c}{Methods} & \multicolumn{5}{c}{Seen Fonts} & \multicolumn{5}{c}{Unseen Fonts} \\ \cmidrule(r){1-4}\cmidrule(r){5-9}\cmidrule(r){10-14}
     P& C& S& N& L1$\downarrow$ & RMSE$\downarrow$  & SSIM $\uparrow$ & LPIPS$\downarrow$ & FID$\downarrow$ & L1$\downarrow$ & RMSE$\downarrow$  & SSIM $\uparrow$ & LPIPS$\downarrow$ & FID$\downarrow$  \\
    \midrule
      &    &    &                       & 
      0.06251 & 0.2105 & 0.7437 & 0.0846 & 17.10 & 0.07841 & 0.2442 & 0.6853 & \underline{0.1198} & 27.98 \\
      \checkmark   &    &    &             & 
      0.06261 & 0.2103 & 0.7434 & 0.0853 & 16.17 & 0.07803 & 0.2435 & 0.6868 & 0.1202 & \underline{26.79} \\
      \checkmark   &    &    & \checkmark              & 
      0.06727 & 0.2221 & 0.7240 & 0.0957 & 17.02 & 0.08009 & 0.2489 & 0.6786 & 0.1259 & 27.12 \\
      \checkmark   &    &    & \checkmark*             & 
      \textbf{0.05952} & \textbf{0.2001} & \textbf{0.7552} & \underline{0.0856} & 23.34 & \underline{0.07519} & \underline{0.2359} & \underline{0.6984} & 0.1224 & 34.83 \\
      \checkmark   &  \checkmark  & &                  & 
      0.06056 & 0.2071 & 0.7506 & 0.0865 & \underline{16.08} & 0.07574 & 0.2399 & 0.6940 & 0.1199 & 27.01 \\
      \checkmark   &  \checkmark  &  \checkmark   &    &  
      \underline{0.05997} & \underline{0.2053} & \underline{0.7538} & \textbf{0.0836} & \textbf{13.13} & \textbf{0.07394} & \textbf{0.2354} & \textbf{0.7007} & \textbf{0.1182} & \textbf{26.51} \\
    \bottomrule
  \end{tabular}
  }
\end{table*}
As Tbl.~\ref{tbl:vs_sota} illustrates, our method outperforms other methods, especially on unseen fonts. 
\wc{DG-Font
leads other comparison methods except on perception metrics. 
But when added our proposed modules, its LPIPS and FID scores get a significant boost and catch up with others both on seen and unseen fonts. Although FUNIT achieves the best FID score on seen fonts, it performs worse on other metrics}.
Fig.~\ref{fig:vs_sota_big} displays the qualitative comparison. Characters generated by ours are of high quality in terms of style consistency and structural correctness. The results of FUNIT, LF-Font, MX-Font, \wc{Fs-Font, and CG-GAN} often have structural errors and incompleteness. 
\wc{Fs-Font select several reference characters from a reference collection through a content-reference mapping, the relationship between a character and its references with common conspicuous components. The reference collection contains around 100 characters and covers almost all components. However, our reference characters are randomly selected and fixed for all source characters, with poor component coverage. Thus, the performance of Fs-Font is not perfectly shown in our few-shot setting.} 
The outputs of DG-Font are great overall but suffer from artifacts and incomplete style transfer.

		

\begin{figure}[t]
	\centering
	\renewcommand\arraystretch{1.45}
	\resizebox{0.85\linewidth}{!}{
	\begin{tabular}{lcc}
		\toprule
		& Seen Fonts & Unseen Fonts \\ \cmidrule(r){1-1} \cmidrule(r){2-2} \cmidrule(r){3-3}
        \makebox[0.11\linewidth][l]{Source} &   \multirow{6}*{\includegraphics[width=0.375\linewidth]{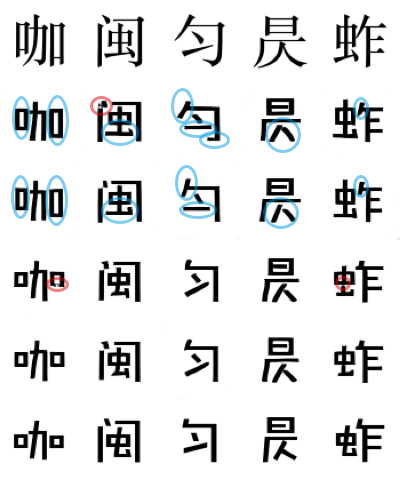}} & \multirow{6}*{\includegraphics[width=0.375\linewidth]{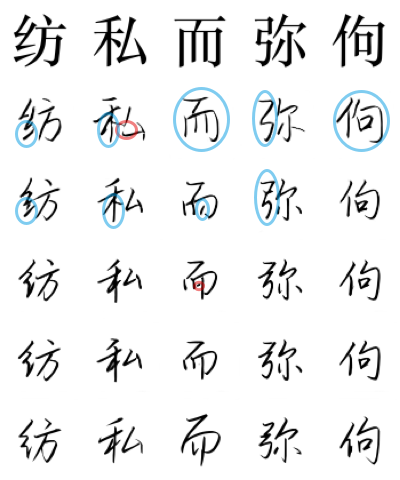}}\\
        \makebox[0.11\linewidth][l]{Baseline} & \\
		\makebox[0.11\linewidth][l]{+P} & \\
        \makebox[0.11\linewidth][l]{+PC} & \\
        \makebox[0.11\linewidth][l]{+PCS} & \\
        \makebox[0.11\linewidth][l]{Target} & \\
        \bottomrule
		
	\end{tabular}
	}
    \vspace{-5pt}
    \caption{Qualitative results in the ablation on different components. P, C, and S are the same notations as Tbl.~\ref{tbl.wcs}. We mark erroneous skeletons with red circles and other mismatch styles with blue circles.}
    \vspace{-10pt}
	\label{fig:vs_wcs}
\end{figure}

\paragraph{User study.} \crc{We conduct a user study to further compare our model with other methods.
We randomly selected 40 font styles~(30 seen fonts and 10 unseen fonts) from the test set, and for each style, 5 test characters were randomly selected. Corresponding character images are generated with our method and the other 6 comparison methods. 20 participants who use Chinese characters every day are asked to pick the best group~(5 character images yielded by one method) for one test style.
Here, the order of these groups is randomly shuffled and we allow multiple choices since the participants might think several synthesized characters are of comparable quality. The results of user study are shown in the last column of \ref{tbl:vs_sota}, which present that our CF-Font gains the highest user preference 21.58\%, surpassing the second place CG-GAN 16.67\% by a large margin.}

\begin{figure}[t]
  \begin{center}
  \vspace{-10pt}
    \includegraphics[width=0.3\textwidth]{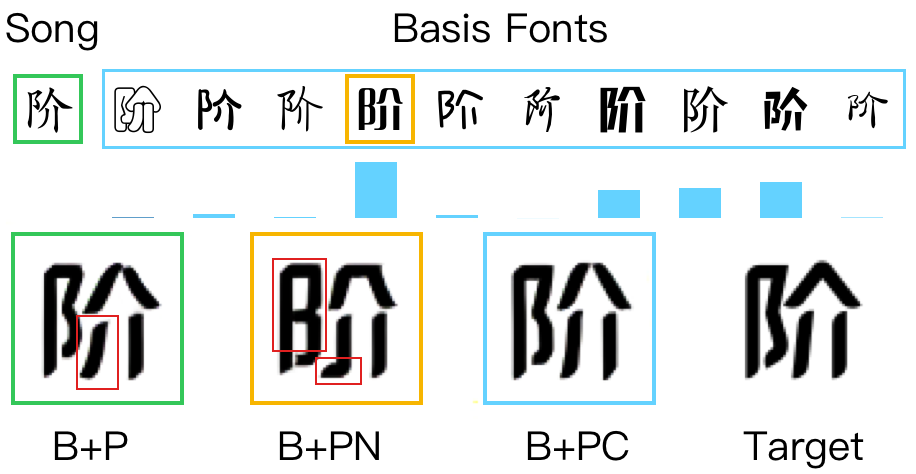}
  \end{center}
  \vspace{-20pt}
  \caption{Comparison between content fusion and retrieval strategy. B represents the baseline (DG-Font), and other notations are the same as Tbl.~\ref{tbl.wcs}.}
  \label{fig:nearest_basis}
\vspace{-10pt}
\end{figure}

\subsection{Ablation Studies}\label{Ablation Studies}

\wc{This subsection shows the effects of all proposed components and discusses how CFM and PCL work in font generation.}
\paragraph{Effectiveness of different components.}
We separate the proposed modules and sequentially add them to DGN to observe the effects of each. The quantitative results can be seen in Tbl.~\ref{tbl.wcs}, \wc{verifying} that PCL, CFM, and ISR all can help improve the quality of generated images. These modules bring not only a numerical improvement but also \wc{a noticeable}
improvement in the visual aspect of geometric structures and stylistic strokes, as displayed in Fig.~\ref{fig:vs_wcs}. In the fourth-to-last line, PCL shows its ability to improve character semantics and skeletons. 
\wc{Moreover,} CFM makes the generated results a big step closer to the target in human perception. In the penultimate line, ISR further refines the detail of results by enhancing the stylistic representation.

\begin{figure}[t]
    \centering
 	\includegraphics[width=\linewidth]{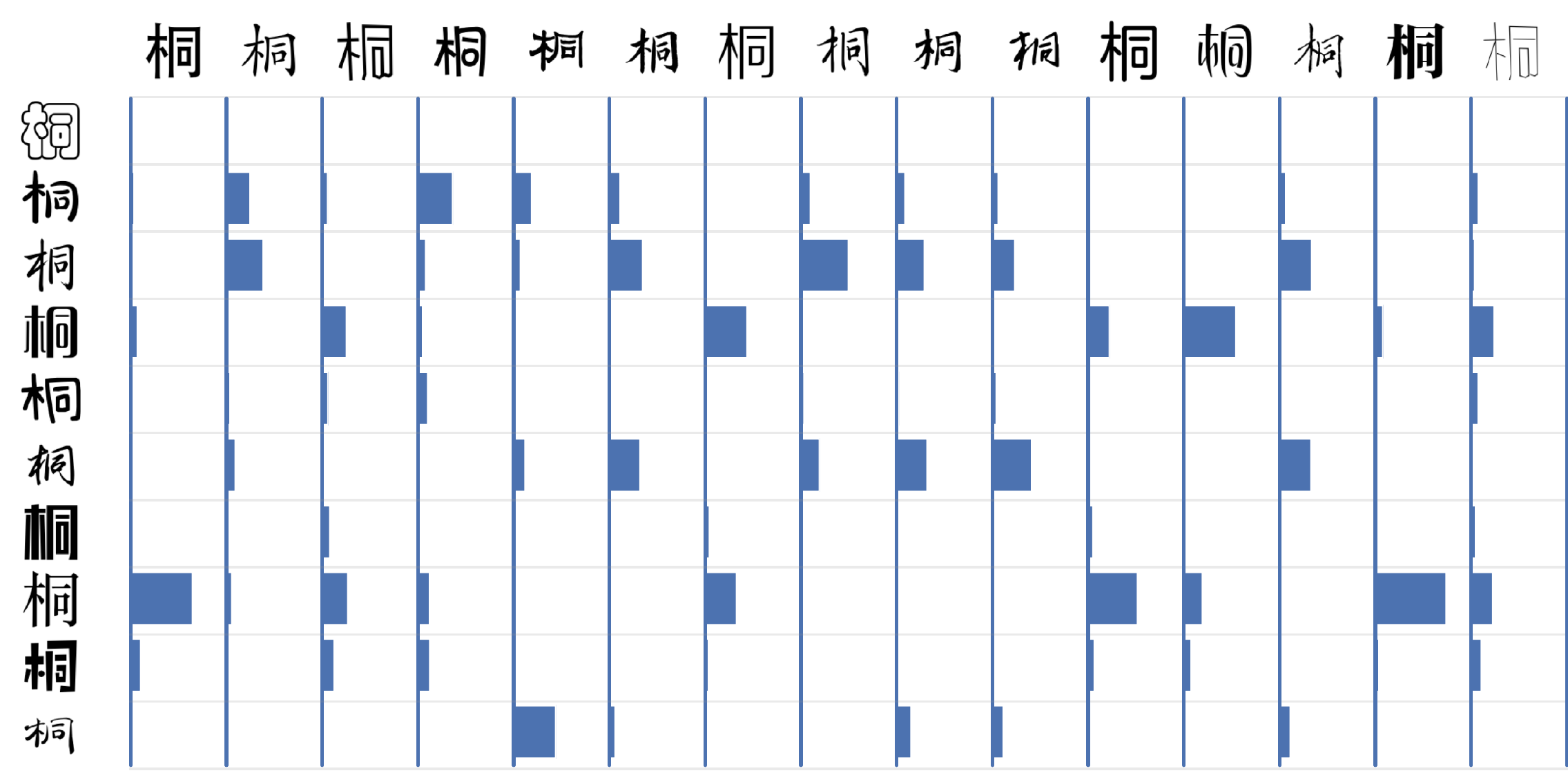}
    \vspace{-15pt}
    \caption{Visualization of weights on basis fonts. We take the character ``Tong" for example. The left column represents the basis fonts, and the top row shows a part of training fonts. The weight on basis fonts of one training font are displayed as a vertical histogram.}
    \vspace{-5pt}
    \label{fig:vis_cf}
\end{figure}

\paragraph{Comparison between content fusion and retrieval strategy.}

Among these modules, CFM is the most efficient one. 
We further analyze where the gain of CFM comes from through a comparison with the retrieval strategy, \ie during the test, we select the closet font for every target font as input from basis fonts and the whole training set (except the target font itself, \ie 239/240 fonts in total for each seen/unseen target font) respectively. The quantitative result is shown in the second to fourth row from the bottom of Tbl.~\ref{tbl.wcs}. It indicates that the result of inputting the closet basis font is much worse than that of content fusion, or even worse than the baseline (using a stand font \emph{Song} all the time). 
\wc{Meanwhile, retrieving the closest font from the whole set gets a comparable results with CF-Font on seen fonts, but not good as it on unseen fonts and FID metrics.}
As Fig.~\ref{fig:nearest_basis} illustrates, the closet font may still be very different on the character skeleton from the target one and will introduce some noises (parts mismatched to the target skeleton). With these observations, we claim that content fusion matters rather than retrieving a close font in CFM.

\paragraph{Variations of PCL.} We use two variations of PCL, PC-WDL, and PC-KL, to train a model respectively. Tbl.~\ref{tbl:pcl} shows the result on unseen fonts and demonstrates that not only PC-WDL, 
PC-KL can also improve the network performance. 
\wc{PC-KL and PC-WDL have similar improvements on pixel-level metrics, but PC-WDL has obvious advantages in FID while PC-KL performs better on LPIPS.}
We attribute this to that benefit from character projection, both of the distribution distance metrics can focus on the global properties, such as skeleton topology. 

\subsection{Evaluation of Basis Selection.}
We visualize the basis fonts and \wc{the corresponding} weights of content fusion here. Taking the character ``Tong" as an example, in Fig.~\ref{fig:vis_cf}, ten images of basis fonts are shown in the left column, fifteen target images with randomly selected fonts are listed in the top row, and the weights of content fusion are plotted in the form of a vertical histogram. We can observe that (a) the basis fonts selected by clustering are indeed visually different from each other (they also can be chosen manually), which means that they are capable of building a space for content fusion; (b) the greater the weight value, the corresponding basis font is more similar to the target font and this proves that content fusion is physically meaningful; (c) the values of these weights are scattered rather than concentrated in a particular basis font, which can also be a reason why the retrieval strategy fails as described in subsection~\ref{Ablation Studies}.

\begin{table}[t]
  \caption{Quantitative evaluation using variations of PCL.}
  \label{tbl:pcl}
  \centering
  \resizebox{0.95\linewidth}{!}{
  \begin{tabular}{llllll}
    \toprule
    method & L1 $\downarrow$ & RMSE $\downarrow$  & SSIM $\uparrow$ & LPIPS $\downarrow$ & FID $\downarrow$ \\
    \midrule
    Baseline & 0.07841 & 0.2442 & 0.6853 & 0.1198 & 27.98 \\
    +PC-KL   & \textbf{0.07802} & \textbf{0.2434} & \textbf{0.6872} & \textbf{0.1191} & 27.72 \\
    +PC-WDL  & 0.07803 & 0.2435 & 0.6868 & 0.1202 & \textbf{26.79} \\
    \bottomrule
  \end{tabular}
  }
\end{table}

\begin{figure}
    \small
    \centering
    \resizebox{0.7\linewidth}{!}{
	\begin{tabular}{l}
		\toprule
		\ \ Source \ \ FUNIT \ LF-Font \ MX-Font \ Fs-Font \ CG-GAN \ DG-Font \ CF-Font \ Target \\
		\midrule
        \includegraphics[width=1.3\linewidth]{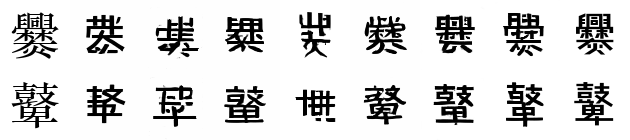} \\
        \bottomrule
	\end{tabular}
	}
\vspace{-5pt}
\caption{Failure case.}\label{fig:failure}
\vspace{-15pt}
\end{figure} 
\subsection{Failure Cases and Limitations}
Fig.~\ref{fig:failure} illustrates a case of generated images of complex characters with many strokes and a tight layout. Although our method works relatively well, many structural errors appear in the first row and some strokes are missed in the second row.
\section{Conclusion}
In this paper, we design a content fusion module and a projected character loss to improve the quality of skeleton transfer in few-shot font generation. We also propose a iterative style-vector refinement strategy to find a better font-level style representation. Experiments demonstrate that our method can outperform existing state-of-the-art methods, and each of the proposed novel modules is effective. 

In the future, we may try vector font generation because vector characters are scale-invariant and more convenient for practical applications. It would be interesting to investigate whether the content fusion strategy can help solve the problem of complex vector font generation. 
\section*{Acknowledgments}
\crc{We thank the anonymous reviewers for their constructive comments. This paper is supported by Information Technology Center and State Key Lab of CAD\&CG, Zhejiang University.}


{\small
\bibliographystyle{ieee_fullname}
\bibliography{egbib}
}

\end{document}